\documentclass[conference]{IEEEtran}
\IEEEoverridecommandlockouts

\usepackage{cite}
\usepackage{amsmath,amssymb,amsfonts}
\usepackage{algorithmic}
\usepackage{graphicx}
\usepackage[font=footnotesize,skip=4pt]{caption}
\usepackage{subcaption} 
\usepackage{float} 
\usepackage{textcomp}
\usepackage{xcolor}
\def\BibTeX{{\rm B\kern-.05em{\sc i\kern-.025em b}\kern-.08em
    T\kern-.1667em\lower.7ex\hbox{E}\kern-.125emX}}

\usepackage{booktabs} 
\usepackage{array}  
\usepackage{tabularx}

\usepackage{tikz}
\usetikzlibrary{shapes.geometric, arrows, positioning, fit, backgrounds, calc}

\pgfdeclarelayer{background}
\pgfsetlayers{background,main}

\tikzset{
  kgarrow/.style={->,>=stealth, line cap=round} 
}

\newcolumntype{T}{>{\ttfamily\raggedright\arraybackslash}X} 
\newcolumntype{Y}{>{\raggedright\arraybackslash}X}          

\begin{document}

\title{A Knowledge-Graph Translation Layer for Mission-Aware Multi-Agent Path Planning in Spatiotemporal Dynamics
}

\author{\IEEEauthorblockN{Edward Holmberg}
\IEEEauthorblockA{\textit{Cannizaro-Livingston Gulf States} \\
\textit{Center for Environmental Informatics}\\
New Orleans, LA, USA \\
eholmber@uno.edu}
\and
\IEEEauthorblockN{Elias Ioup}
\IEEEauthorblockA{\textit{Center for Geospatial Sciences} \\
\textit{Naval Research Laboratory}\\
Stennis Space Center, Mississippi, USA \\
elias.z.ioup.civ@us.navy.mil}

\and
\IEEEauthorblockN{Mahdi Abdelguerfi}
\IEEEauthorblockA{\textit{Cannizaro-Livingston Gulf States} \\
\textit{Center for Environmental Informatics}\\
New Orleans, LA, USA \\
gulfsceidirector@uno.edu}
}

\maketitle

\begin{abstract}
The coordination of autonomous agents in dynamic environments is hampered by the semantic gap between high-level mission objectives and low-level planner inputs. To address this, we introduce a framework centered on a Knowledge Graph (KG) that functions as an intelligent translation layer. The KG's two-plane architecture compiles declarative facts into per-agent, mission-aware ``worldviews" and physics-aware traversal rules, decoupling mission semantics from a domain-agnostic planner. This allows complex, coordinated paths to be modified simply by changing facts in the KG. A case study involving Autonomous Underwater Vehicles (AUVs) in the Gulf of Mexico visually demonstrates the end-to-end process and quantitatively proves that different declarative policies produce distinct, high-performing outcomes. This work establishes the KG not merely as a data repository, but as a powerful, stateful orchestrator for creating adaptive and explainable autonomous systems.
\end{abstract}


\section{Introduction}

The effective coordination of autonomous agents, be they robotic vehicles, sensor networks, or even human teams, in dynamic, real-world environments presents a formidable challenge.\cite{lavalle2006planning,mei2024multi_agent} Mission success often depends on the ability to translate a complex set of domain-specific considerations into actionable plans. These considerations include evolving mission goals, operational policies, agent capabilities, and a constant stream of spatiotemporal data from a changing environment. This translation process is further complicated by the need to optimize resource allocation, ensure responsiveness to unforeseen events, and account for the stochastic nature of the world.\cite{wu2014path_problems,mei2024multi_agent}

Traditional planning systems have struggled to bridge this gap between high-level, semantic mission context and the low-level inputs required by optimization algorithms. Many current approaches rely on static, brittle configuration files and a cumbersome, human-in-the-loop workflow. This paradigm is ill-suited for dynamic environments; any change in mission parameters or environmental conditions often requires manual intervention and a full, computationally expensive replan, making it difficult to adapt in a timely and stable manner. This highlights a critical need for a more stateful, semantic, and incremental substrate for planning.

To address these limitations, our framework is built around a Knowledge Graph (KG) that acts as an intelligent \textbf{translation layer}.  The KG is responsible for maintaining a rich, semantic model of the mission's spatiotemporal context, including environmental dynamics, agent capabilities, and operational policies. It then compiles this knowledge into a set of universal, planner-agnostic artifacts, mission-aware tensors and dynamic seam rules, that are consumed by a generic path planner. This separation of concerns allows the planner to remain domain-agnostic, while the KG provides the mission-specific intelligence.

This architecture enables a powerful, adaptive planning loop. The KG not only provides the initial context but also orchestrates the entire process, facilitating the coordination of long-horizon temporal paths. Crucially, it supports reactive replanning by identifying which parts of the plan are affected by real-time updates and instructing the planner to recompute only what is necessary.  We demonstrate the power of this approach through a case study involving a fleet of Autonomous Underwater Vehicle (AUV) gliders in the dynamic ocean environment of the Gulf of Mexico, extending earlier visualization-based forecasting work \cite{holmberg2014data_visualization}  and early distributed geospatial architectures \cite{chung2001gids}. However, the framework itself is designed to be general and applicable to a wide range of multi-agent coordination problems. \cite{rudnick2015ocean_research,chassignet2007hycom}

The primary contributions of this work are:

\begin{itemize}
    \item A formal \textbf{Knowledge-Graph translation layer} with two compilation maps: a data plane ($\Phi_2$) that generates per-agent, mission-aware heatmaps, and a control plane ($\Phi_1$) that produces dynamic seam rules and rewards. \cite{khan2018knowledge, tenorth2013knowrob,kostavelis2015semanticmaps}

    \item A method for generating \textbf{per-agent, mission-aware tensors} that encode policies, constraints, and environmental factors, allowing different agents to have unique ``world views'' without changing the planner.

    \item A demonstration of \textbf{adaptive planning} through an MPC-style loop, where the KG's ability to perform incremental recompilation of ``dirty windows'' enables fast and stable replans in response to mid-mission perturbations. \cite{garcia1989mpc,mayne2014mpc}

\end{itemize}

\section{Background}

\subsection{Spatiotemporal Points of Interest (POIs)}
A Point of Interest (POI) is a location in both space and time where data collection or observation is of high value. The criteria for defining a POI are flexible and practitioner-defined, allowing the concept to be adapted to a wide range of missions and scientific objectives \cite{Holmberg2024ROBUST,holmberg2022geo_spatiotemporal}.

\subsection{Dynamic Environmental Conditions}
In many real-world scenarios, the environment is not static. Factors such as ocean currents, wind patterns, or even urban traffic flows change over time, altering the cost and feasibility of traversing different paths. For low-power autonomous agents, such as AUV gliders, these dynamic conditions are not just a minor inconvenience; they are a dominant factor that dictates the energy, time, and safety costs of a mission \cite{rudnick2015ocean_research, chassignet2007hycom}. Any effective planning system must be able to account for and adapt to these evolving conditions. \cite{zermelo1931,nmpc_ocean_paths_lolla2014}

\subsection{Multi-Agent Coordination}
The challenge of path planning is compounded when multiple agents are involved. Multi-agent coordination requires not only finding optimal paths for each individual agent but also ensuring that the collective behavior of the team is efficient and deconflicted. The objective is to maximize overall mission success, for example, by ensuring comprehensive data coverage of all significant POIs, while minimizing redundant effort and avoiding inter-agent collisions \cite{mei2024multi_agent, sharon2015cbs}.

\subsection{Knowledge Graphs for Representing Complex Systems}
Knowledge Graphs (KGs) are a powerful tool for representing and reasoning over complex, heterogeneous information.By modeling entities and their relationships in a structured graph, KGs can capture 
the rich semantics of a domain \cite{ladner2012mining}. In the context of path planning, a KG can encode not just the physical layout of an environment but also the temporal dynamics, the capabilities and policies of different agents, and the high-level constraints of a mission. This provides a formal, machine-readable representation of the mission context that can be queried and updated in real-time \cite{khan2018knowledge, tenorth2013knowrob,kostavelis2015semanticmaps}.

\section{Related Work}

Our framework integrates concepts from three distinct but related fields: path planning in dynamic environments, multi-agent coordination, and the application of knowledge graphs to robotics.

\subsection{Path Planning in Dynamic Environments}
Path planning in environments with time-varying conditions is a well-established field. Classical approaches often rely on graph-based algorithms like Dijkstra's or A* applied to a time-expanded graph, where each node represents a location at a specific time \cite{lavalle2006planning, wu2014path_problems}. To account for dynamics like ocean currents, methods such as Zermelo's navigation problem have been incorporated to find time-optimal paths \cite{zermelo1931, nmpc_ocean_paths_lolla2014}. More recently, Model Predictive Control (MPC) and other receding-horizon techniques have been used to enable reactive replanning by repeatedly solving short-horizon problems \cite{garcia1989mpc,mayne2014mpc}.

Spatial network query optimizations such as the AKNN M-tree approach \cite{Ioup2007AKNN} inform our efficient waypoint sampling within each planning window. While powerful, these methods often assume that the objectives and constraints of the mission are static and can be encoded directly into the cost function. Our work complements these approaches by introducing a \textbf{Knowledge Graph as a translation layer}, which allows the mission's high-level, semantic context, including policies, uncertainty, and evolving scientific priorities, to be dynamically compiled into the planner's inputs.

\subsection{Multi-Agent Coordination and Task Allocation}
The coordination of multiple agents, often referred to as Multi-Agent Path Finding (MAPF), is another rich area of research \cite{Holmberg2024ROBUST,holmberg2022geo_spatiotemporal}. Techniques range from centralized approaches that find globally optimal, collision-free paths, such as Conflict-Based Search (CBS) \cite{sharon2015cbs, Holmberg2024WAITR}, to decentralized methods that are more scalable but may not guarantee optimality. For task allocation in dynamic environments, market-based mechanisms like the Consensus-Based Bundle Algorithm (CBBA) have proven effective for assigning tasks to agents in a distributed manner \cite{choi2009cbba, bertsekas1992auction, holmberg2023stroobnet}.

Our framework addresses multi-agent coordination not by proposing a new search algorithm but by providing a mechanism for generating \textbf{per-agent, policy-aware "world views."} By compiling different mission-aware heatmaps for each agent based on their assigned policies, our KG enables a team of heterogeneous agents to pursue diverse objectives within a unified planning structure.

\subsection{Knowledge Graphs in Robotics and Planning}
The use of knowledge graphs and semantic ontologies in robotics is a growing field. Much of the existing work has focused on creating "semantic maps" that enrich a robot's understanding of its environment, for example, by labeling objects and rooms to enable more complex task planning \cite{tenorth2013knowrob,kostavelis2015semanticmaps}. Other approaches have used KGs to inform reinforcement learning policies or to provide a basis for explainable AI in autonomous systems \cite{khan2018knowledge}.

Our work is distinct in that it does not use the KG as a direct representation of the search space. Instead, we formalize the KG's role as a \textbf{translation layer} with two explicit compilation maps: a data plane ($\Phi_2$) that synthesizes mission-aware tensors and a control plane ($\Phi_1$) that generates dynamic seam rules. To our knowledge, this use of a KG to declaratively compile the inputs for a windowed, stitched spatiotemporal planner is a novel contribution.

\section{Methodology}


\begin{figure*}[!t]
  \centering
  \resizebox{\textwidth}{!}{%
  \begin{tikzpicture}[
    node distance=1.4cm and 2.0cm,
    input/.style={rectangle, rounded corners, draw, minimum height=2.2em, text width=12em, align=center},
    plane/.style={rectangle, draw, minimum height=3.6em, text width=12em, align=center, fill=white},
    artifact/.style={rectangle, draw, dashed, minimum height=3.0em, text width=16em, align=center, fill=white},
    plan/.style={rectangle, rounded corners, draw, minimum height=2.2em, text width=12em, align=center, fill=white},
    sheet/.style={rectangle, draw, minimum height=2.2em, text width=10.2em, align=center, fill=white, inner ysep=4pt},
    flow/.style={->,>=stealth, line cap=round, thick},
    feed/.style={->,>=stealth, line cap=round, thick, dashed},
    kgbox/.style={rectangle, rounded corners, very thick, fill=none}
  ]

  \node[font=\large\bfseries] (inputsTitle) {Inputs};
  \node (ontologies) [input, below=6mm of inputsTitle] {Domain Ontologies \&\\ Models};
  \node (rtupdates)  [input, below=8mm of ontologies] {Real-time Updates\\ (Forecast Refresh)};

  \node (dataplane)    [plane, right=3.0cm of ontologies] {Data Plane};
  \node (controlplane) [plane, below=1.3cm of dataplane]  {Control Plane};

  \node (kg) [kgbox, draw=none, fit=(dataplane)(controlplane)] {};

  \coordinate (dp_w_low) at ($(dataplane.west)+(0,-6mm)$);
  \draw[flow] (ontologies.east) -- (dataplane.west);
  \draw[flow] (rtupdates.east)  -- ++(6mm,0) |- (dp_w_low);

  \draw[feed] (controlplane.north) --
    node[pos=0.5, right=0mm, align=center, font=\small, fill=white, inner sep=1pt]
      {Rejected Paths /\\ Feedback}
    (dataplane.south);

  \coordinate (S) at ($(kg.east)+(3.4cm,0)$);
  \node (sheetF) [sheet] at ($(S)+(-14mm, 14mm)$) {Feasibility};
  \node (sheetP) [sheet] at ($(S)+(-9mm,   9mm)$) {Policy};
  \node (sheetU) [sheet] at ($(S)+(-4mm,   4mm)$) {Uncertainty};
  \node (sheetR) [sheet] at ($(S)+( 0mm,   0mm)$) {Reward};

  \node (stack_at_data_y) at ($(sheetR.east |- dataplane.center)$) {};
  \node (agg)     [plane, minimum height=2.2em, text width=10em, right=2.0cm of stack_at_data_y]
                  {Aggregate /\\ Mission Tensor};
  \node (planner) [plane, right=1.6cm of agg] {Agnostic Path Planner};

  \coordinate (handoff) at ($(sheetR.east |- dataplane.center)+(6mm,0)$);
  \draw[flow] (handoff) -- (agg.west);

  \draw[flow] (agg.east) -- (planner.west);

  \node (finalplan) [plan, right=2.2cm of controlplane] {Coordinated\\ Mission Plan};

  \node (selector)  [plan, minimum height=2.2em, text width=10em]
        at ($(agg.center |- finalplan.center)$) {Selector /\\ Coordinator};

  \node (pathlets)  [artifact]
        at ($(planner.center |- finalplan.center)$)
        {Pathlet Candidates\\ \small (All Potential paths for each Agent\\ across all Time Windows)};

  \draw[flow] (controlplane.east) -- (finalplan.west);
  \draw[flow] (selector.west)    -- (finalplan.east);
  \draw[flow] (selector.east)     -- (pathlets.west);

  \draw[flow] (planner.south)     -- (pathlets.north);

  \coordinate (KG_left_x)  at ($(dataplane.west)$);
  \coordinate (KG_right_x) at ($(agg.east)$);           
  \coordinate (KG_NW) at ($(KG_left_x |- dataplane.north)+(-8pt, 10pt)$);
  \coordinate (KG_SW) at ($(KG_left_x |- controlplane.south)+(-8pt,-10pt)$);
  \coordinate (KG_NE) at ($(KG_right_x |- dataplane.north)+(10pt, 10pt)$);
  \coordinate (KG_SE) at ($(KG_right_x |- controlplane.south)+(10pt,-10pt)$);

  \draw[very thick, rounded corners] (KG_NW) rectangle (KG_SE);
  \node[font=\bfseries, fill=white, inner sep=1pt]
  at ($ (KG_NW)!0.5!(KG_NE) $) {Knowledge Graph};

  \end{tikzpicture}%
  }
  \caption{Architectural overview of the Knowledge Graph as a central orchestrator. The \textbf{Data Plane ($\Phi_2$)} compiles semantic inputs into a stack of numerical \textit{Mission Tensors}, which are fused into an \textit{Aggregate Tensor} for the \textbf{Agnostic Path Planner}. The planner generates \textit{Pathlet Candidates}. The \textbf{Control Plane ($\Phi_1$)} then orchestrates the selection and coordination from these candidates to produce the final \textit{Coordinated Mission Plan}. Feedback from rejected paths can inform subsequent compilations, creating an adaptive planning loop.}
  \label{fig:kg_pipeline}
\end{figure*}
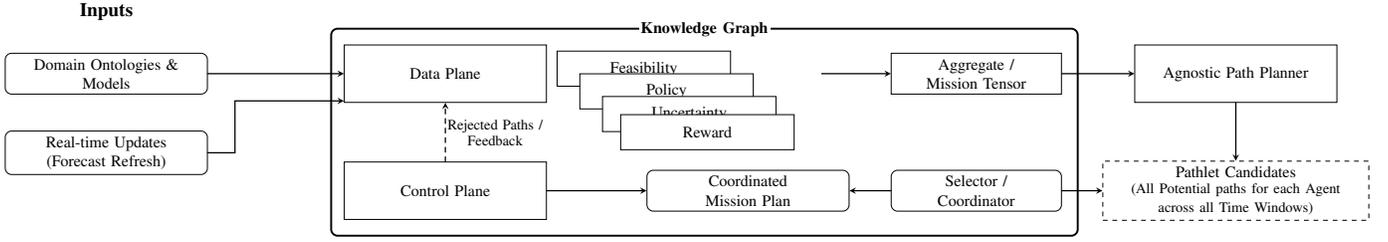

Our framework is designed around a central Knowledge Graph (KG) that acts as a translation layer, separating mission-specific knowledge from a domain-agnostic planner. The KG ingests high-level mission context and compiles it into two sets of artifacts that the planner consumes: a series of per-agent, per-window mission-aware heatmaps, and a set of dynamic seam rules for stitching paths across time. This architecture is illustrated in Figure \ref{fig:kg_pipeline}.

\subsection{The Knowledge Graph Model}

To formalize the mission context, we model it as a typed, temporal KG. The schema defines the core entities (``nouns'') and their relationships (``verbs''), and is aligned with established standards like OWL-Time, GeoSPARQL, and PROV-O \cite{w3c_owl_time,ogc_geosparql,prov_o}. This allows mission parameters to be managed as declarative facts. The schema is detailed in two parts: Table \ref{tab:kg_nouns} describes the core classes, and Table \ref{tab:kg_verbs} defines their key relational properties.

\begin{table}[t]
\centering
\caption{Core KG Schema: Representative Classes (Nouns)}
\label{tab:kg_nouns}
\renewcommand{\arraystretch}{1.08}
\setlength{\tabcolsep}{4pt}
\scriptsize
\begin{tabular}{@{}>{\ttfamily}p{0.35\columnwidth} >{\raggedright\arraybackslash}p{0.61\columnwidth}@{}}
\toprule
\textbf{Class} & \textbf{Description} \\
\midrule
ex:TimeWindow       & A temporal interval representing a single planning step, annotated with forecast confidence. \\
ex:ValueLayer        & A raw or derived scientific data raster (e.g., SST frontness) for a specific time window \cite{cayula1992sied,belkin2009fronts}. \\
ex:Constraint       & A spatial restriction, such as a no-go zone or a soft-penalty area, with a defined geometry. \\
ex:Policy           & A named set of weights and rules that declaratively define an agent's behavior and objectives. \\
ex:Agent            & An autonomous entity with a designated policy and a set of physical capabilities. \\
ex:Event            & A discrete, high-value mission objective, such as a point of interest to be sampled. \\
ex:TensorArtifact  & A compiled, mission-aware tensor; an output of the Data Plane ($\Phi_2$). \\
ex:NavGraph         & A traversable waypoint-and-edge graph for a time window, generated from a heatmap. \\
\bottomrule
\end{tabular}
\end{table}

\begin{table}[t]
\centering
\caption{Core KG Schema: Key Properties (Verbs)}
\label{tab:kg_verbs}
\renewcommand{\arraystretch}{1.05}
\setlength{\tabcolsep}{3.5pt}
\scriptsize
\begin{tabular}{@{}>{\ttfamily}p{0.30\columnwidth}
                >{\ttfamily}p{0.30\columnwidth}
                >{\raggedright\arraybackslash}p{0.36\columnwidth}@{}}
\toprule
\textbf{Property} & \textbf{(Domain $\rightarrow$ Range)} & \textbf{Description} \\
\midrule
ex:forWindow        & (ex:BL, ex:CF $\rightarrow$ ex:TW)      & Links environmental data to a time window. \\
ex:usesPolicy       & (ex:Ag $\rightarrow$ ex:Pol)            & Assigns a behavioral policy to an agent. \\
ex:hasCapabilities  & (ex:Ag $\rightarrow$ ex:AC)             & Links an agent to its physical parameters. \\
geo:asWKT           & (ex:Con $\rightarrow$ xsd:string)       & Provides the geometry of a constraint. \\
prov:wasDerivedFrom & (ex:TA $\rightarrow$ ex:BL, ex:Pol)     & Traces an artifact to its source facts. \\
ex:from / ex:to     & (ex:TE $\rightarrow$ ex:WP)             & Defines a directed edge between waypoints. \\
ex:costHours        & (ex:EC $\rightarrow$ xsd:double)        & Stores the computed traversal time for an edge. \\
\bottomrule
\multicolumn{3}{@{}p{\columnwidth}@{}}{\footnotesize
\textit{Abbreviations}: Ag (Agent), AC (AgentCapabilities), BL (BaseLayer), CF (CurrentField), Con (Constraint), EC (EdgeCost), TA (TensorArtifact), Pol (Policy), TE (TraverseEdge), TW (TimeWindow), WP (Waypoint).}
\end{tabular}
\end{table}

\subsection{The Two-Plane Translation Architecture}

The KG's primary role is to act as a compiler through two distinct but coordinated compilation maps: a Data Plane ($\Phi_2$) and a Control Plane ($\Phi_1$).

\subsubsection{The Data Plane ($\Phi_2$): Compiling Mission Tensors}

The Data Plane is responsible for translating high-level semantic context into a stack of per-agent, per-window numerical rasters, which we formalize as \textbf{Mission Tensors}. These tensors represent an agent's specific ``worldview" at a given time. The compilation for agent $a$ at time window $t$ is a multi-step process:

\begin{enumerate}
    \item \textbf{Layer Blending}: The compiler queries the KG for the active \texttt{ex:ValueLayer} ($B_t$), any relevant \texttt{ex:PriorLayer}s ($P^{(a)}_{t,k}$), and the weights ($\alpha, \beta, \gamma$) from the agent's assigned \texttt{ex:Policy}. These are blended to form a preliminary value field. In our AUV example, this might combine SST ``frontness" with a prior for staying over a continental shelf.
    \item \textbf{Constraint Application}: Active \texttt{ex:Constraint}s are queried. Geometries of \texttt{no\_go} constraints are used to create a hard mask ($\mathcal{M}_{\text{hard}}$), while \texttt{soft} constraints produce an attenuation mask ($\mathcal{M}_{\text{soft}}$).
    \item \textbf{Confidence Scaling}: The resulting field is scaled by the \texttt{ex:confidence} ($w_t$) of the \texttt{ex:TimeWindow}, de-weighting the value of less certain future states.
    \item \textbf{Aggregation}: These individual tensors; Reward, Uncertainty, Policy Priors, and Feasibility; are fused into an \textbf{Aggregate Mission Tensor} ($U_t^{(a)}$), a single utility field that serves as the primary input to the path planner. The fusion is defined by:

     \begin{equation}
    \label{eq:fusion} 
    \begin{split}
        U_{t}^{(a)} = \mathcal{N}\Big( w_t \cdot \Big[ &\alpha^{(a)}B_t + \sum_k \beta^{(a)}_k P^{(a)}_{t,k} + \gamma^{(a)}\mathcal{F}(B_t) \Big] \\
        &\cdot \mathcal{M}_{\text{soft}} \Big) \odot \mathcal{M}_{\text{hard}}
    \end{split}
    \end{equation}
        where $\mathcal{N}$ is a normalization function and $\mathcal{F}$ represents an optional transformation like gradient calculation for frontness.
\end{enumerate}

\subsubsection{The Control Plane ($\Phi_1$): Governing Coordination and Selection}

The Control Plane provides the logic and rules for coordination and decision-making. Unlike the Data Plane, which generates dense tensors, the Control Plane provides functions and discrete facts used by the Selector/Coordinator module.

\begin{itemize}
    \item \textbf{Seam Rules and Costs}: The Control Plane generates feasibility functions ($F_{t \to t+1}$) and scoring functions ($S_{t \to t+1}$) for traversing between waypoints. Feasibility considers agent capabilities (e.g., \texttt{baseTravel}, \texttt{boostGain}) and environmental factors (\texttt{WorkField}), while the scoring function uses the policy's $\lambda$-weights to compute a cost based on time, energy, and hazard exposure.
    \item \textbf{Coordination Logic}: It provides the rules for deconfliction. This includes enforcing \texttt{ex:Event} capacity, applying minimum-separation penalties between agents, and implementing fairness policies (e.g., scarcity-aware tie-breaking).
    \item \textbf{Feedback Loop}: When the Selector rejects a set of paths due to a conflict, the Control Plane is responsible for generating "cooling" constraints or updating priors (e.g., `ex:softOverrides`) that are fed back to the Data Plane. This informs the next compilation cycle, discouraging agents from repeatedly proposing conflicting plans.
\end{itemize}

\subsection{Agnostic Planning and Coordinated Selection}

This module consumes the artifacts generated by the KG to produce the final, coordinated plan. It consists of two stages: candidate generation and selection.

\subsubsection{Pathlet Candidate Generation}
The Agnostic Path Planner is a domain-unaware optimization engine. Its task is to generate a diverse set of high-quality, long-horizon path options, or "Pathlet Candidates," for each agent.
\begin{enumerate}
    \item \textbf{Waypoint Sampling}: For each window, high-value waypoints are extracted from the Aggregate Mission Tensor ($U_t^{(a)}$).
    \item \textbf{Temporal Path Stitching}: A Viterbi-style dynamic programming algorithm or similar graph search method is used to find multiple high-scoring paths across the full time horizon \cite{viterbi1967, forney1973viterbi}. This process is guided entirely by the seam rules and costs provided by the Control Plane ($\Phi_1$). It produces a set of potential end-to-end paths for each agent \cite{holmberg2023stroobnet}.
\end{enumerate}

\subsubsection{Coordinated Plan Selection}
The Selector/Coordinator module is the embodiment of the Control Plane's logic. It takes the set of all Pathlet Candidates and produces a single, deconflicted mission plan. It does this by:
\begin{enumerate}
    \item \textbf{Applying Hard Constraints}: It rejects any candidate paths that violate mission-critical constraints (e.g., event capacities).
    \item \textbf{Resolving Conflicts}: It uses a greedy assignment or auction mechanism to resolve spatial conflicts, where multiple agents propose paths to the same or nearby locations. Tie-breaking can be guided by control-plane policies, such as prioritizing the agent with fewer alternative high-value options (scarcity) \cite{choi2009cbba,bertsekas1992auction}.
    \item \textbf{Committing the Plan}: The winning set of paths constitutes the `Coordinated Mission Plan`. Assignments are written back to the KG as `ex:Assignment` facts for auditing and to inform future planning cycles.
\end{enumerate}

\subsection{Architectural Benefits of the KG Approach}

This architecture provides four key capabilities that are difficult to achieve with traditional static configurations.

\paragraph{Provenance and Explainability} Every compiled artifact and planning decision can be traced back to its source facts in the KG. A ``why-trace" can explain why a path was chosen by citing the specific policy weights, environmental data, and constraints that made it optimal.

\paragraph{Reasoning and Validation} The KG is not just a data store; it can apply rules. For example, a SHACL profile validates data integrity (e.g., $0 \le w_t \le 1$), while domain-specific rules can alter planning outcomes (e.g., escalating a `soft` constraint to `hard` when a ``safe" policy is active) \cite{w3c_shacl, w3c_sparql11}.

\paragraph{Queryability} The use of SPARQL allows operators to perform ``what-if" analyses and audits, such as asking, ``Which planned paths would become infeasible if an agent's energy budget were reduced by 30\%?"

\paragraph{Incrementality and Replanning} The KG's structure enables efficient, low-latency updates. When a fact is changed, the KG can identify the ``dirty" windows and recompile only the necessary artifacts, allowing the planner to quickly and stably adapt to new information in an MPC-style loop.

\subsubsection{Core Schema Entities (The ``Nouns")}

The KG is built around a set of core, general-purpose classes that can be adapted to any multi-agent domain. The main entities include:

\begin{itemize}
    \item \textbf{\texttt{ex:TimeWindow}}: Represents a temporal interval for planning. It is aligned with OWL-Time and is annotated with an \texttt{ex:confidence} score ($w_t \in [0,1]$) to represent, for instance, the decaying certainty of a weather or ocean forecast over time.

    \item \textbf{\texttt{ex:ValueLayer}}: Represents the primary source of mission value for a time window. This abstract layer can model any objective-driven data. In our AUV case study, this is derived from Sea Surface Temperature (SST) gradients to identify oceanic fronts (``frontness").

    \item \textbf{\texttt{ex:WorkLayer}}: Represents a time-varying vector field of external forces that affect the cost and feasibility of traversal. This generalizes environmental effects; for our AUV gliders, this class encapsulates HYCOM ocean current data.

    \item \textbf{\texttt{ex:Constraint}}: Aligned with GeoSPARQL, this defines a spatial restriction. Each constraint has a \texttt{ex:kind} and a geometry (\texttt{geo:asWKT}). Examples include \texttt{`no\_go`} zones for landmasses and `soft` penalty zones for marine sanctuaries.

    \item \textbf{\texttt{ex:Policy}}: A named set of declarative weights and rules that define an agent's behavior. This allows for heterogeneous objectives; for example, one AUV might use a "fuel-efficient" policy with a high $\lambda_{\text{energy}}$, while another uses a "fastest" policy with a high $\lambda_{\text{time}}$.

    \item \textbf{\texttt{ex:Agent}}: Represents an autonomous entity, such as a robot or a vehicle, with a designated policy (\texttt{ex:usesPolicy}) and physical capabilities (\texttt{ex:hasCapabilities}). In our experiments, these are modeled as AUV gliders with specific speed and energy parameters.
\end{itemize}

\subsubsection{Data Ingestion and Provenance (The ``Verbs")}
Data is ingested into the KG through a set of adapters that populate these classes. For example, HYCOM NetCDF files are processed to create `:CurrentField` and `:BaseLayer` instances for each time window. All data is tagged with provenance information using PROV-O, allowing for full traceability of the inputs to the planning process.

\subsection{The Translation Layer: Compilation Maps}
The KG's primary function is to translate its stored knowledge into a format the planner can use. This is achieved through two compilation maps.

\subsubsection{Data Plane ($\Phi_2$): KG to Mission-Aware Heatmaps}
The $\Phi_2$ map is responsible for generating a mission-aware heatmap, $H_{t}^{*(a)}$, for each agent `a` and time window `t`. This heatmap represents the value of sampling at each location in the grid. The compilation is a multi-step process:
\begin{enumerate}
    \item \textbf{Blending:} A base scientific layer ($B_t$) is blended with agent-specific priors and an optional "frontness" emphasis, controlled by weights ($\alpha, \beta, \gamma$) from the agent's policy.
    \item \textbf{Applying Constraints:} The heatmap is then modified by applying the relevant constraints from the KG. Hard constraints (e.g., land) zero out the corresponding cells, while soft constraints apply an attenuation factor.
    \item \textbf{Scaling by Confidence:} The entire heatmap is scaled by the window's confidence score ($w_t$), down-weighting the value of less certain future states.
    \item \textbf{Normalization:} The final heatmap is normalized to a range of [0,1].
\end{enumerate}

\subsubsection{Control Plane ($\Phi_1$): KG to Stitching Rules and Rewards}
The $\Phi_1$ map provides the rules for connecting path segments across time windows:
\begin{itemize}
    \item \textbf{Node Rewards ($R_t$):} The value of visiting a waypoint (derived from the heatmap) is scaled by the window confidence ($w_t$).
    \item \textbf{Seam Feasibility ($F_{t \to t+1}$):} A seam between two waypoints is deemed feasible only if it does not violate any hard constraints and is achievable within the agent's travel budget, considering the assistance or resistance from ocean currents.
    \item \textbf{Seam Score ($S_{t \to t+1}$):} The cost of traversing a feasible seam is calculated as a weighted sum of travel time, energy, hazard exposure, and an uncertainty penalty, with the weights ($\lambda$ values) drawn from the agent's policy.
\end{itemize}
\subsection{The Planner Architecture}
The planner is designed to be domain-agnostic and operates in three stages:

\subsubsection{Per-Window Micro-Path Generation}
For each time window, the planner uses the mission-aware heatmap ($H_t^*$) to identify a set of high-value waypoints. It then generates a set of optimal "micro-paths" within that window. This process is highly parallelizable.

\subsubsection{Stitching Across Windows}
The planner then uses a Viterbi-style dynamic programming algorithm to ``stitch" the micro-paths together into a globally coherent, long-horizon plan. This stitching process is guided by the seam feasibility and score functions provided by the KG's $\Phi_1$ map.

\subsubsection{MPC and Replanning}
The framework supports a Model Predictive Control (MPC) loop. When the KG is updated with new information, it identifies the "dirty" windows affected by the change. The planner then only needs to recompile the artifacts and re-stitch the plan for those specific windows, enabling fast and stable replanning.


\subsection{The Knowledge Graph Architecture: A Declarative Mission Orchestrator}

The core of our framework is a temporal, semantic Knowledge Graph (KG) that functions as an intelligent \textbf{translation layer and mission orchestrator}. It is not merely a database or a configuration file; it is an active, stateful model of the mission's entire context. Its primary purpose is to bridge the \textit{``semantic-to-numeric gap"} by compiling high-level, declarative knowledge into the precise, low-level numerical inputs required by a domain-agnostic path planner.

\subsubsection{Core Architectural Principles}
\begin{enumerate}
    \item \textbf{Declarative Mission Modeling}: All aspects of the mission, from agent capabilities and policy objectives to environmental constraints and scientific goals, are represented as explicit, typed entities and relationships (facts). To change the mission, an operator changes the facts, not the code.
    \item \textbf{Decoupling of Concerns}: The KG cleanly separates the ``what" and ``why" of the mission (the semantic context) from the ``how" of pathfinding (the numerical optimization). The planner remains domain-agnostic; all mission intelligence is injected via the compiled artifacts.
    \item \textbf{End-to-End Provenance}: Every artifact produced by the KG and every decision made by the planner can be traced back to the specific set of facts that informed it. This provides a complete audit trail for explainability, debugging, and mission analysis.
    \item \textbf{Dynamic Re-planning and Incrementality}: The KG is designed for dynamic environments. It can efficiently identify the downstream effects of a change (e.g., an updated current forecast) and trigger a re-compilation of only the affected artifacts, enabling fast, stable, and incremental replanning.
\end{enumerate}

\subsubsection{The Ontological Foundation (The Schema)}
The KG's schema provides the formal vocabulary for describing the mission. It is aligned with established standards to ensure semantic interoperability and a rigorous foundation.

\paragraph{Namespaces}
We utilize the following standard and custom namespaces:
\begin{description}
  \item[\texttt{ex:}] Core mission ontology.
  \item[\texttt{prov:}] PROV-O for provenance.
  \item[\texttt{time:}] OWL-Time for temporal entities.
  \item[\texttt{geo:}] GeoSPARQL for spatial data.
  \item[\texttt{ssn:}] Sensor/observation (SSN/SOSA).
\end{description}

\paragraph{High-Level Class Hierarchy}
\begin{figure}[t]
  \centering
  \scriptsize 
  \setlength{\baselineskip}{0.94\baselineskip} 
  \fbox{%
    \begin{minipage}{0.94\columnwidth}
      \ttfamily
owl:Thing\\
|-- \textbf{ex:MissionEntity} \quad (mission elements)\\
|   |-- ex:Agent\\
|   |-- ex:Policy\\
|   |-- ex:Mission\\
|   \texttt{\`--} ex:Event\\
|-- \textbf{ex:SpatiotemporalEntity} \quad (entities with space/time)\\
|   |-- time:TemporalEntity $\rightarrow$ ex:TimeWindow\\
|   |-- ex:GridSpec\\
|   |-- ex:WorkField\\
|   |-- ex:ValueLayer\\
|   \texttt{\`--} ex:Constraint\\
|-- \textbf{ex:Artifact} \quad (compiled outputs of the KG)\\
|   |-- ex:TensorArtifact\\
|   |-- ex:NavGraph\\
|   |-- ex:Waypoint\\
|   |-- ex:TraverseEdge\\
|   \texttt{\`--} ex:EdgeCost\\
\texttt{\`--} \textbf{prov:Activity} \quad (actions: compilation/planning)\\
    |-- ex:CompilePhi2\\
    \texttt{\`--} ex:PlanRun
    \end{minipage}%
  }
  \caption{High-level class hierarchy (ASCII tree).}
  \label{fig:kg_class_hierarchy}
\end{figure}

\subsubsection{Mission Context: The Populated Graph (Nouns \& Verbs)}
The KG is populated with instances of the schema classes, creating a rich, interconnected graph of the mission's state organized into logical planes.

\paragraph{Spatiotemporal Plane (The World)}
This plane describes the physical and temporal grid of the operational environment.

\begin{table}[t]
  \centering
  \caption{Spatiotemporal Plane Schema}
  \label{tab:schema_spatiotemporal}
  \begin{tabularx}{\columnwidth}{@{}T Y T@{}}
    \toprule
    \textbf{Class} & \textbf{Description} & \textbf{Key Properties (Verbs)} \\
    \midrule
    ex:TimeWindow   & A discrete time slice of the mission. &
      ex:index (int), time:hasBeginning (xsd:dateTime), ex:confidence (double) \\
    ex:GridSpec     & Defines the raster grid for all spatial data. &
      ex:rows (int), ex:cols (int), geo:asWKT (bbox), ex:pixelSizeKm (double) \\
    ex:CurrentField & Ocean current vector fields for a window. &
      ex:forWindow ($\to$ ex:TimeWindow), ex:gridRefU (string), ssn:observedProperty \\
    ex:BaseLayer    & A raw or derived scientific field. &
      ex:forWindow ($\to$ ex:TimeWindow), ex:kind ("frontness"), prov:wasDerivedFrom \\
    \bottomrule
  \end{tabularx}
\end{table}

\paragraph{Operational \& Policy Plane (The Rules)}
This plane defines the mission's objectives, rules, and constraints.

\begin{table}[t]
\centering
\caption{Operational \& Policy Plane Schema}
\label{tab:schema_policy}
\scriptsize
\setlength{\tabcolsep}{3pt}
\renewcommand{\arraystretch}{1.05}
\begin{tabularx}{\columnwidth}{@{}T Y T@{}}
\toprule
\textbf{Class} & \textbf{Description} & \textbf{Key Properties (Verbs)} \\
\midrule
ex:Constraint & A spatial restriction. &
  ex:kind (``no\_go''/``soft''), geo:asWKT (polygon), ex:attenuation (double), ex:appliesIn \\
ex:Policy & Named behavioral parameters. &
  ex:alpha\_base, ex:lambda\_time, ex:priors (JSON), ex:softOverrides (JSON) \\
ex:PriorLayer & Spatial preference map (e.g., corridor). &
  ex:name (string), ex:gridRef (string) \\
ex:Event & Discrete, high-value mission occurrence. &
  ex:forWindow, geo:asWKT, ex:value, ex:capacity, ex:expiresAfter \\
\bottomrule
\end{tabularx}
\end{table}

\paragraph{Agent \& Platform Plane (The Actors)}
This plane describes the agents and their capabilities.

\begin{table}[t]
\centering
\caption{Agent \& Platform Plane Schema}
\label{tab:schema_agent}
\scriptsize
\setlength{\tabcolsep}{3pt}
\renewcommand{\arraystretch}{1.05}
\begin{tabularx}{\columnwidth}{@{}T Y T@{}}
\toprule
\textbf{Class} & \textbf{Description} & \textbf{Key Properties (Verbs)} \\
\midrule
ex:Agent & Autonomous entity executing missions. &
  ex:id, ex:mobility, ex:usesPolicy ($\rightarrow$ ex:Policy), ex:hasCapabilities \\
ex:AgentCapabilities & Physical limits of an agent. &
  ex:cruiseSpeedKts, ex:maxSpeedKts, ex:energyPerKm, ex:boostGain \\
ex:AgentState & Agent state at a specific time. &
  ex:forAgent, ex:forWindow, geo:asWKT (position), ex:headingDeg \\
\bottomrule
\end{tabularx}
\end{table}

\paragraph{Execution \& Provenance Plane (The Artifacts and Audit Trail)}
This plane stores the compiled outputs, linking them back to their inputs.


\begin{table}[t]
\centering
\caption{Execution \& Provenance Plane Schema}
\label{tab:schema_provenance}
\scriptsize
\setlength{\tabcolsep}{3pt}
\renewcommand{\arraystretch}{1.05}
\begin{tabularx}{\columnwidth}{@{}T Y T@{}}
\toprule
\textbf{Class} & \textbf{Description} & \textbf{Key Properties (Verbs)} \\
\midrule
ex:HeatmapArtifact & Compiled, mission-aware heatmap. &
ex:forAgent, ex:forWindow, ex:hash, prov:wasDerivedFrom \\
ex:NavGraph & Traversable waypoint graph for a window. &
ex:forWindow, ex:movementMode ("hops"/"edges") \\
ex:Waypoint & Node in the NavGraph. &
ex:inGraph, geo:asWKT, ex:score (double) \\
ex:TraverseEdge & Edge in the NavGraph. &
ex:inGraph, ex:from, ex:to \\
ex:EdgeCost & Physics-aware costs per agent/edge. &
ex:forEdge, ex:forAgent, ex:costHours, ex:costEnergy, ex:costRisk \\
ex:PlanRun & A planning execution instance. &
ex:forWindow, ex:worldVersion, ex:status ("ok","stale") \\
ex:Assignment & Final decision for an agent/window. &
ex:forAgent, ex:forWindow, ex:waypointWKT, prov:wasGeneratedBy \\
\bottomrule
\end{tabularx}
\end{table}

\subsubsection{The Compilation Engine: The Two-Plane Translation Layer}
The KG's core function is to execute two compilation maps, \textbf{\(\Phi_2\) (Data Plane)} and \textbf{\(\Phi_1\) (Control Plane)}, which translate the semantic graph into numerical artifacts and rules for the planner.

\paragraph{Data Plane (\(\Phi_2\)): KG to Mission-Aware Heatmaps}
This map generates a mission-aware heatmap, \(H^{*}\), for each agent and time window through a series of graph queries and raster operations:
\begin{enumerate}
  \item \textbf{Query for Inputs}: For a given \texttt{(agent, t)}, query the KG to retrieve the agent's \texttt{Policy}, the \texttt{TimeWindow}'s \texttt{confidence}, active \texttt{Constraints}, and relevant \texttt{BaseLayer} and \texttt{PriorLayer} URIs.
  \item \textbf{Blend Layers}: Retrieve raster data for the layers and blend them according to the \texttt{alpha}, \texttt{beta}, and \texttt{gamma} weights in the agent's \texttt{Policy}.
  \item \textbf{Apply Constraints}: Apply masks for active \texttt{soft} and \texttt{no\_go} constraints, attenuating or zeroing out values as specified.
  \item \textbf{Scale and Normalize}: Scale the result by the window's confidence score and normalize to \([0,1]\).
  \item \textbf{Materialize Artifact}: Create a \texttt{TensorArtifact} node in the KG, linking it via \texttt{wasDerivedFrom} to all KG entities used in its creation.
\end{enumerate}

\paragraph{Control Plane (\(\Phi_1\)): KG to Traversal Rules and Costs}
This map generates the rules for moving between locations.
\begin{enumerate}
  \item \textbf{Generate \texttt{NavGraph}}: For each \texttt{(agent, t)}, sample a set of \texttt{Waypoints} from its \(H^{*}\) and link them with \texttt{TraverseEdge}s to form a \texttt{NavGraph}.
  \item \textbf{Annotate \texttt{EdgeCost}}: For each \texttt{(edge, agent, t)}, create an \texttt{ex:EdgeCost} node by querying for agent capabilities, environmental factors (\texttt{WorkField}), risk factors (neighboring waypoint scores and constraint proximity), and policy weights.
  \item \textbf{Provide Feasibility Rules}: The planner queries for feasibility by checking the existence and values of compiled \texttt{EdgeCost} nodes (e.g., ensuring \texttt{costHours} is within the window's duration).
\end{enumerate}

\subsubsection{Operational Aspects}
These features make the KG a dynamic and trustworthy system.

\paragraph{Queryability (SPARQL)}
The KG can be introspected in real time. Examples include:
\begin{itemize}
  \item \textbf{``What-If'' Analysis}: \textit{``Which \texttt{TraverseEdge}s would become too costly for `AUV-Alpha' if we switch to the `safe' policy?''}
  \item \textbf{Provenance Trace}: \textit{``Show all \texttt{ValueLayer}s and \texttt{Constraint}s used to generate this \texttt{TensorArtifact}.''}
  \item \textbf{Mission Audit}: \textit{``List all \texttt{Assignment}s generated by a specific \texttt{PlanRun}.''}
\end{itemize}

\paragraph{Validation and Integrity (SHACL)}
A set of SHACL (Shapes Constraint Language) rules ensures logical consistency, such as range checks on confidence values and correct cardinality between entities (e.g., an agent must use exactly one policy).

\paragraph{Incrementality and Replanning (The MPC Loop)}
The KG's structure is optimized for dynamic updates. When a fact is changed (e.g., an updated current forecast), the KG identifies all ``dirty'' dependent artifacts and triggers a re-compilation of only those specific items, enabling fast and stable replanning.


\section{Experimental Setup}

To evaluate the performance and demonstrate the modularity of our KG-driven planning framework, we conducted a series of evaluations based on a realistic case study: a fleet of autonomous underwater vehicles (AUVs) tasked with a multi-day informative path planning mission in the Gulf of Mexico. Our goal is to scientifically isolate the contributions of the KG's Data Plane ($\Phi_2$) and Control Plane ($\Phi_1$).

\subsection{Data Source and Mission Context}
The environmental data for our experiments were sourced from a series of HYCOM (Hybrid Coordinate Ocean Model) NetCDF files \cite{chassignet2007hycom}. We used a mission horizon covering 7 temporal windows. For each time window, we ingest the following facts into the KG:
\begin{itemize}
    \item \textbf{\texttt{ex:WorkField}}: Surface-level ocean currents (U, V), used by the Control Plane for physics-aware traversal costs.
    \item \textbf{\texttt{ex:ValueLayer}}: A scientific value layer derived from the absolute difference in Sea Surface Temperature (SST) between consecutive days, $|\Delta \mathrm{SST}|$. This proxy for oceanic fronts serves as the base input for the Data Plane.
\end{itemize}

\subsection{Knowledge Graph Configuration}
The KG was configured with entities representing the mission's semantic and physical context:
\begin{itemize}
    \item \textbf{\texttt{ex:TimeWindow}}: The 7-window mission horizon, with forecast confidence modeled using an exponential decay function, $w_t = e^{-0.12t}$.
    \item \textbf{\texttt{ex:Constraint}}: A hard \texttt{no\_go} constraint was created from a land mask derived from the HYCOM data, with a 5-cell buffer. A soft constraint with an attenuation factor of 0.4 represented a protected marine sanctuary. A vertical "science corridor" was also modeled as a positive \texttt{ex:PriorLayer}.
    \item \textbf{\texttt{ex:Agent} \& \texttt{ex:Policy}}: We modeled three agents, each with distinct capabilities and assigned to one of two policies:
        \begin{itemize}
            \item \textbf{Policy ``FAST":} Emphasizes the base scientific value and frontness, with a high `alpha\_base` and `gamma\_front`. Corresponds to your ``fastest" agent concept.
            \item \textbf{Policy ``SAFE":} Has a lower weight on frontness and a stronger attenuation penalty for entering the soft constraint (sanctuary) zone. Corresponds to your ``safe" agent.
        \end{itemize}
\end{itemize}

\subsection{Planner and Selector Configuration}
The domain-agnostic planner and selector modules operate as follows:
\begin{itemize}
    \item \textbf{Waypoint Generation}: We use a Proximal Recurrence (PR) algorithm implemented via convolution to identify a set of up to 20 high-value waypoints (or cluster centers) per time window from the mission-aware heatmap \cite{holmberg2023stroobnet, Holmberg2024WAITR, Holmberg2024ROBUST}. This step is a programmatic implementation of creating an \texttt{ex:NavGraph}.
    \item \textbf{Horizon Path Planning}: A Viterbi-style dynamic programming algorithm stitches a cost-minimal path across the 7-window horizon for each agent, using the traversal costs and feasibility rules from the Control Plane.
    \item \textbf{Team Coordination}: A greedy selection process is used for deconfliction. The highest-scoring agent path is selected, and its nodes are masked (or "cooled") before planning the next agent's path to prevent overlap.
\end{itemize}

\subsection{Evaluation Metrics}
We evaluated the performance of each configuration using the following metrics:
\begin{itemize}
    \item \textbf{Mission Reward}: The cumulative score of the chosen path, calculated as the sum of rewards from the selected waypoints. For a fair comparison across all ablations, the reward is based on the waypoint values derived from the \textit{Full-KG} heatmap.
    \item \textbf{Path Cost / Unique Coverage}: A measure of efficiency, calculated from the number of unique cells covered by the agents' sensor footprints along their paths, rewarding broader, non-redundant exploration.
    \item \textbf{Constraint Adherence}: The number of times a planned path enters a hard constraint (\texttt{no\_go}) zone.
    \item \textbf{Replan Latency}: For perturbation experiments (e.g., adding a new constraint mid-mission), we measure the wall-clock time required to recompile the necessary artifacts and generate a new plan.
\end{itemize}

\section{Results and Analysis}

Our experiments visually and quantitatively demonstrate the benefits of the Knowledge Graph (KG) as a translation layer. The results are presented in a cumulative story: first, we show how the KG compiles semantic data into a mission-aware ``worldview"; second, we illustrate how that worldview is used to generate a complete multi-horizon plan; and finally, we demonstrate the power of using declarative policies to alter agent behavior and mission outcomes.

\subsection{The Data Plane: Compiling a Mission-Aware Worldview}

\begin{figure}[h!]
  \centering
  \begin{subfigure}{0.48\columnwidth}
    \centering\includegraphics[width=\linewidth]{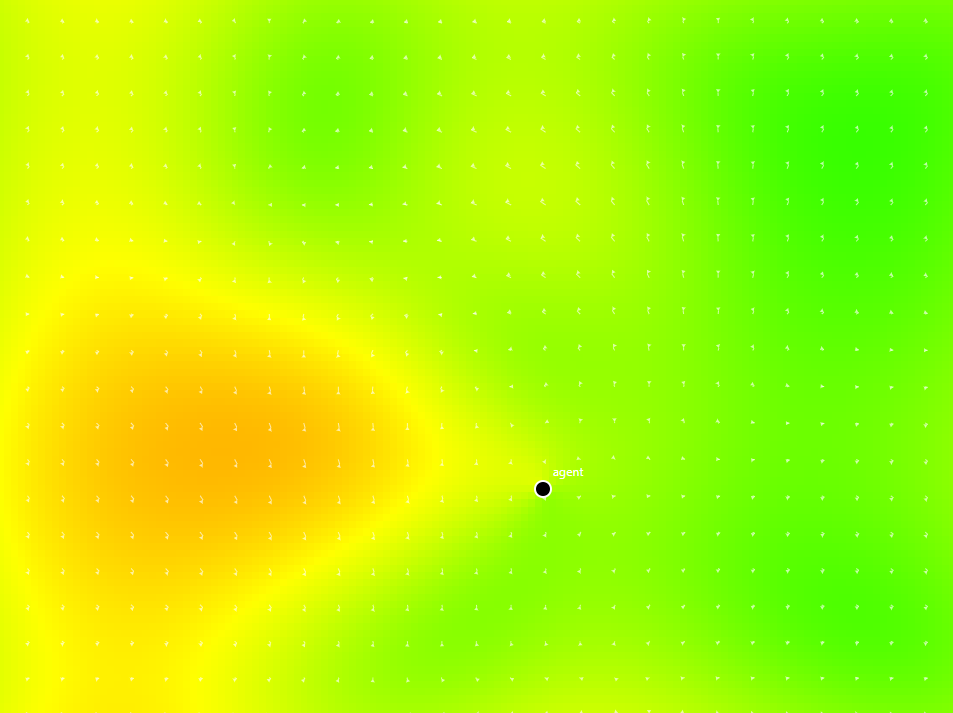}
    \caption{Pose 1}\label{fig:worldview_pose1}
  \end{subfigure}\hfill
  \begin{subfigure}{0.48\columnwidth}
    \centering\includegraphics[width=\linewidth]{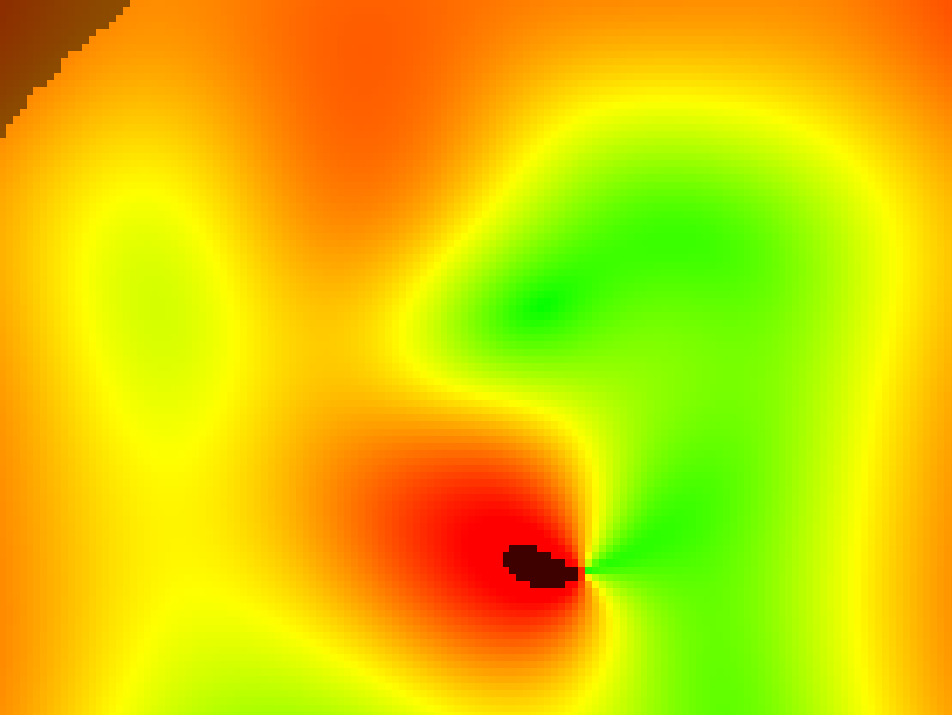}
    \caption{Pose 3}\label{fig:worldview_pose3}
  \end{subfigure}
  \caption{Per-agent work tensors at two poses (green=easy, red=costly).}
  \label{fig:worldview_poses}
\end{figure}

The process begins with the \textbf{Data Plane ($\Phi_2$)}, which translates high-level semantics into low-level planner inputs. The KG enables the creation of per-agent ``work” tensors that define the feasibility and cost of movement by sampling the local vector field (e.g., ocean currents). In Figure~\ref{fig:worldview_poses}, green indicates easy motion, red indicates higher effort, and dark red is infeasible.

\begin{figure}[h!]
  \centering
  \includegraphics[width=\columnwidth]{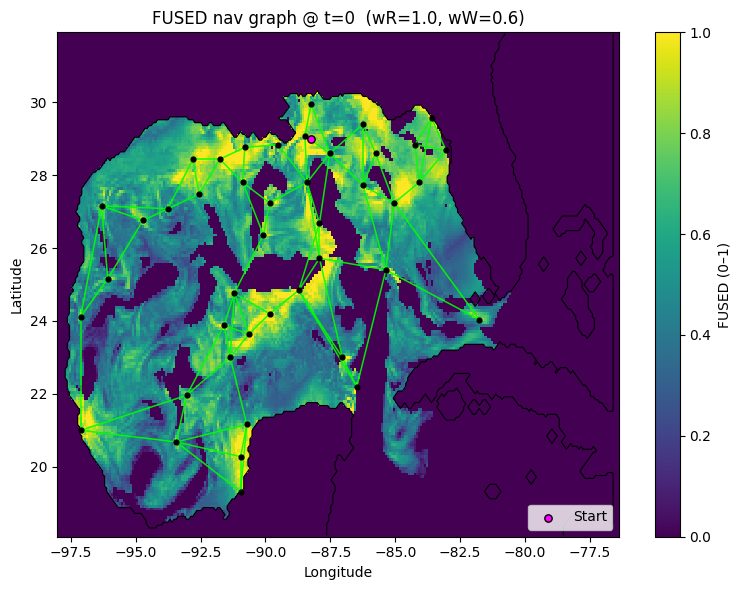}
  \caption{Aggregated mission tensor (\emph{FUSED}) from KG compilation (Data Plane $\Phi_2$).}
  \label{fig:tensor_fusion}
\end{figure}

\begin{figure}[h!]
  \centering
  \includegraphics[width=0.9\columnwidth]{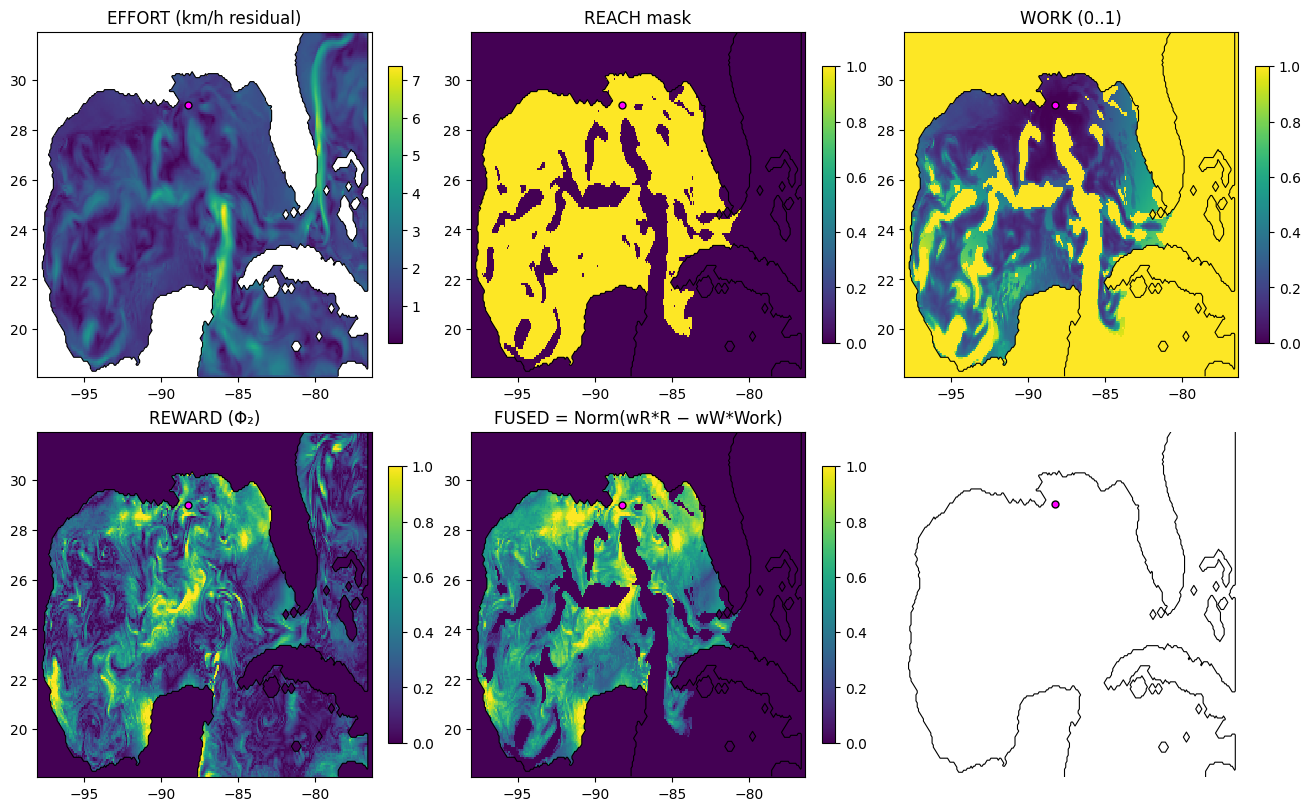}
  \caption{Agent “worldview”: value (colors) with environmental work vectors.}
  \label{fig:agent_worldview}
\end{figure}

These individual tensors: \texttt{REWARD} (from scientific value), \texttt{REACH} (from constraints), and \texttt{WORK} are then fused via the KG into a single, mission-aware \texttt{FUSED} tensor per Eq.~\ref{eq:fusion}. This aggregated \emph{mission tensor}, shown in Figure~\ref{fig:tensor_fusion}, constitutes the agent's complete “worldview” for planning. Figure~\ref{fig:agent_worldview} explicitly overlays the value and work components to show how they co-shape this final view.

\subsection{Multi-Horizon Plan Generation}

\begin{figure}[h!]
  \centering
  \includegraphics[width=0.9\columnwidth]{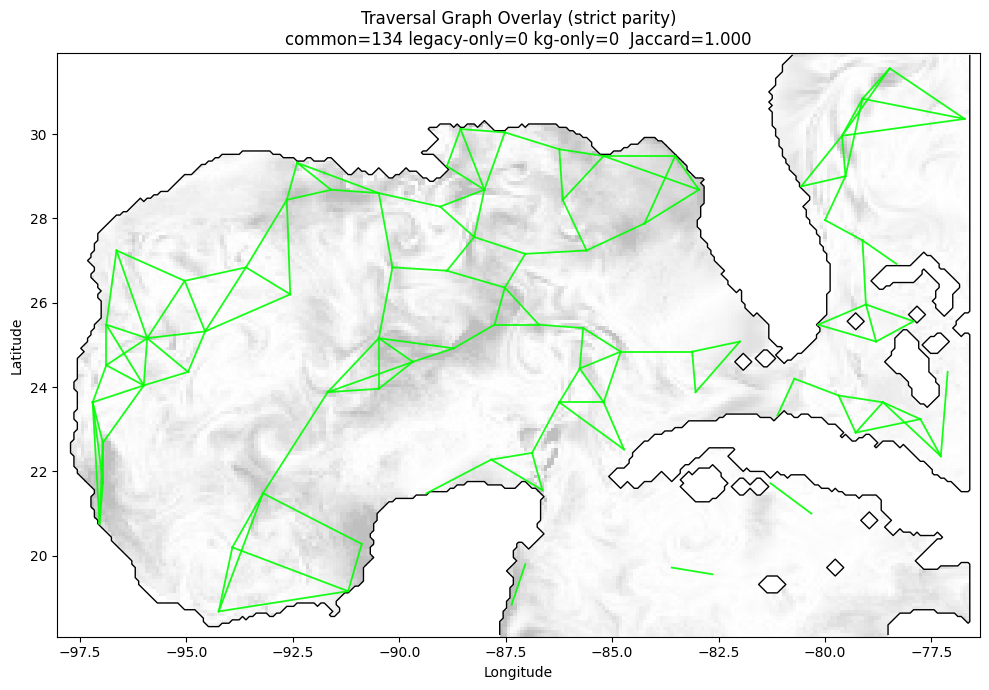}
  \caption{A baseline \texttt{NavGraph} sampled from a naïve, value-only layer.}
  \label{fig:nav_graph_t0}
\end{figure}

Using the compiled artifacts, the planner generates a complete, multi-horizon strategy. We first establish a baseline, shown in Figure~\ref{fig:nav_graph_t0}, where a \texttt{NavGraph} is constructed from a naïve, value-only tensor, consistent with prior work.

\begin{figure}[h!]
  \centering
  \includegraphics[width=0.92\columnwidth]{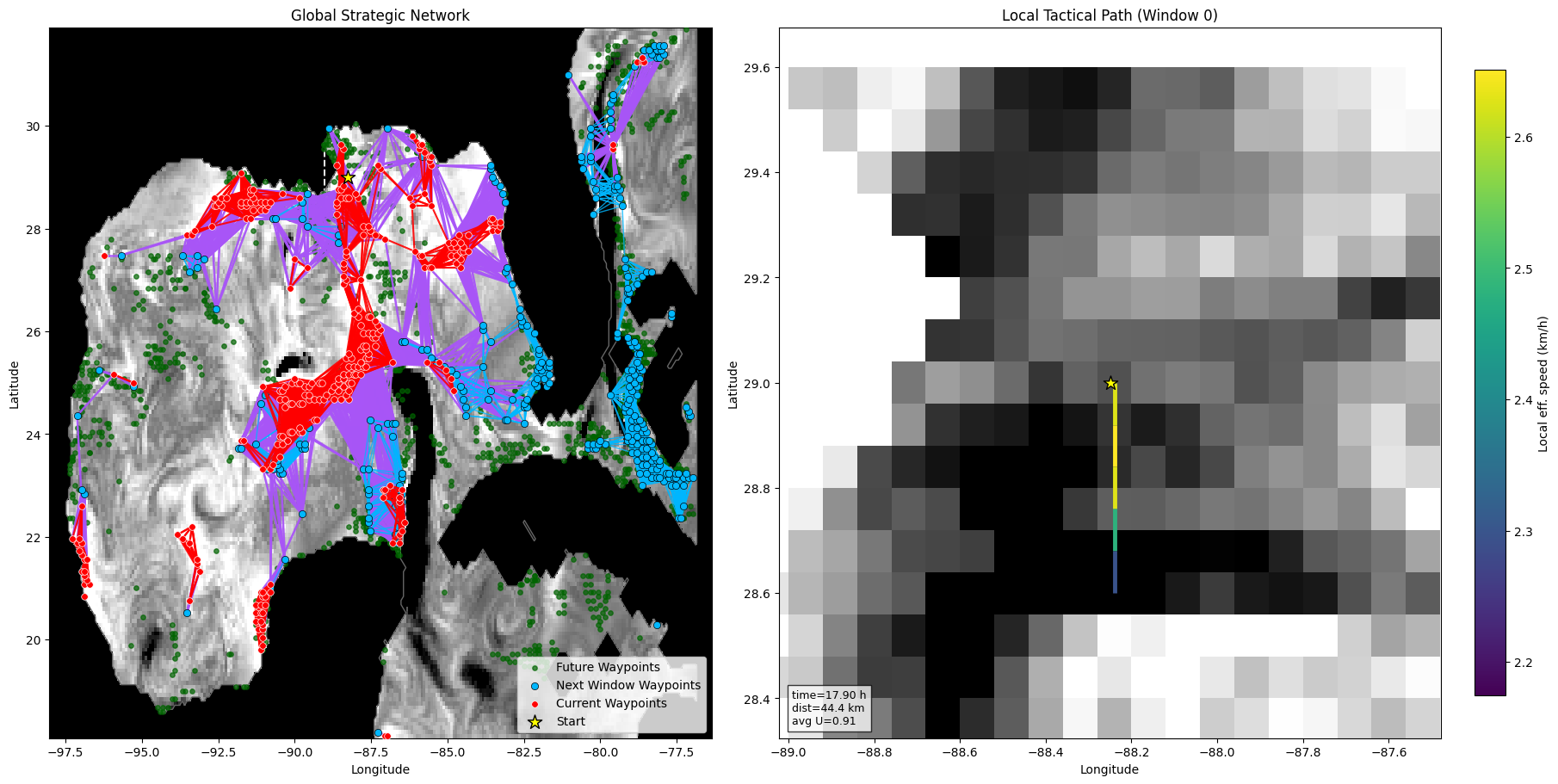}
  \caption{Temporal stitching across time windows with staging (purple) to maximize future payoff.}
  \label{fig:graph_stitching}
\end{figure}

\begin{figure}[h!]
  \centering
  \includegraphics[width=0.92\columnwidth]{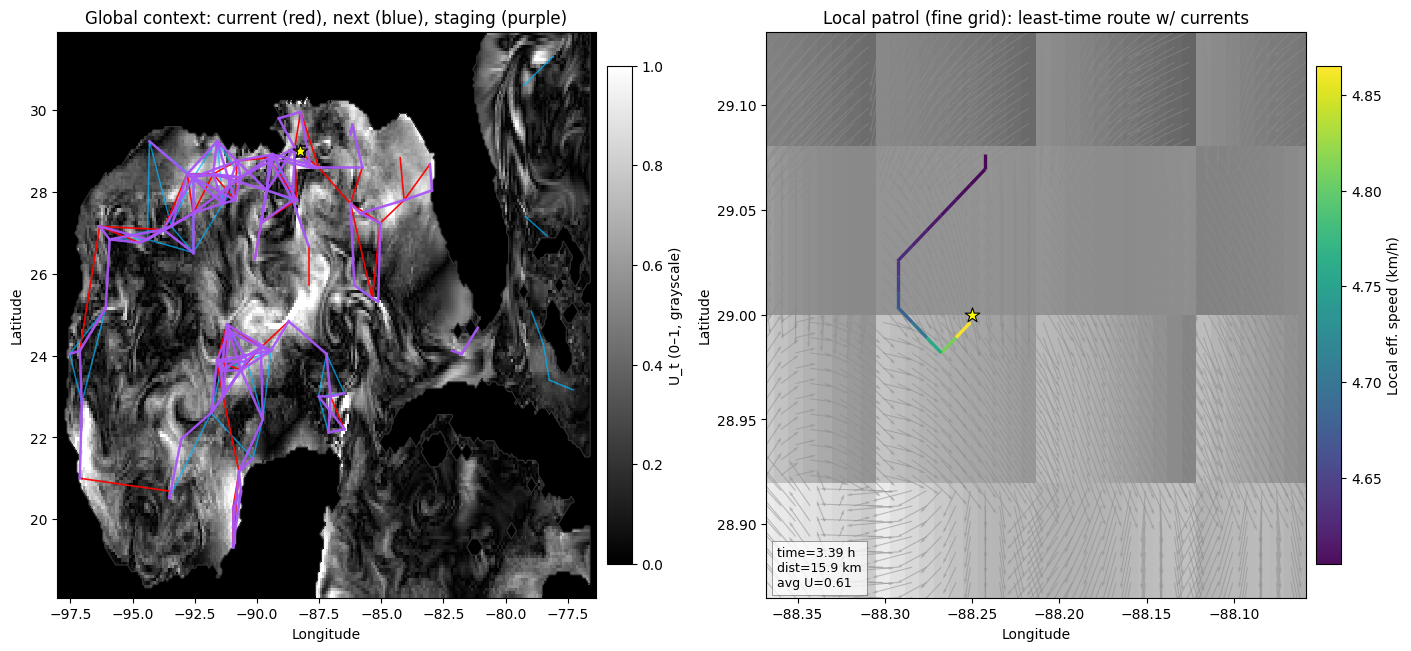}
  \caption{Control Plane ($\Phi_1$): a local path bends to exploit favorable currents.}
  \label{fig:local_path}
\end{figure}

Building upon this, the full framework stitches these per-window graphs across time, guided by the \textbf{Control Plane ($\Phi_1$)}. Figure~\ref{fig:graph_stitching} shows this temporal stitching, where purple "staging" edges position the agent to maximize future opportunities. The Control Plane also computes physics-aware tactical paths between waypoints, bending with currents to minimize travel time (Figure~\ref{fig:local_path}).

\subsection{Declarative Policy-Driven Behavior}

\begin{figure}[h!]
  \centering
  \includegraphics[width=\columnwidth]{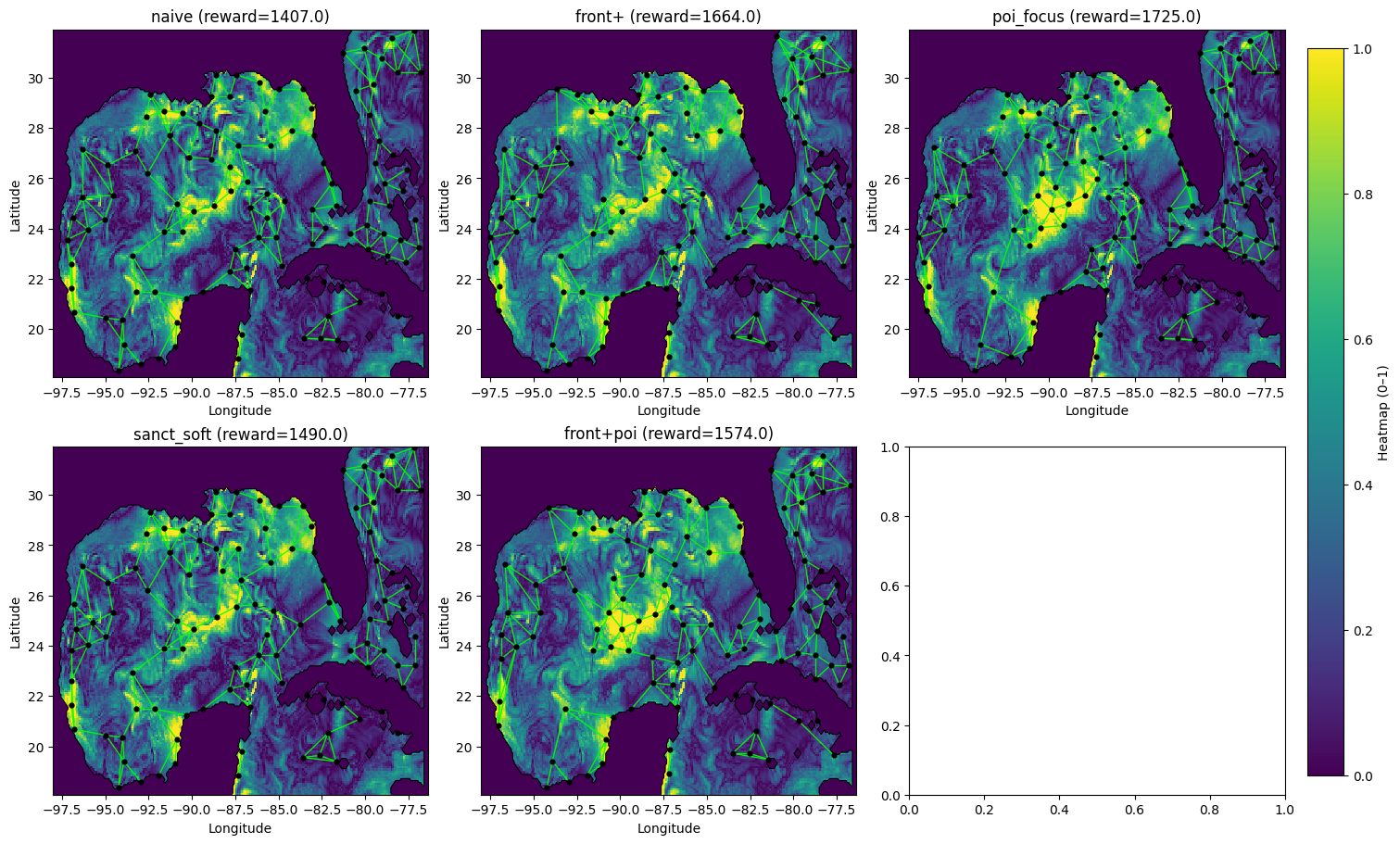}
  \caption{End-to-end effect of policy: aggregated mission tensors (left) and induced \texttt{NavGraphs} (right) change with declarative settings in the KG.}
  \label{fig:global_and_local_plan}
\end{figure}

The framework's primary strength is its ability to generate distinct plans by changing a single \texttt{ex:Policy} fact in the KG. Table~\ref{tab:policy_results} quantifies the mission reward for five different policies. Figure~\ref{fig:global_and_local_plan} provides the key visual evidence, showing how different declarative policy "knobs" in the KG produce qualitatively different mission tensors and, consequently, different induced \texttt{NavGraphs}. This directly links the high-level semantic change to the low-level planner inputs and, ultimately, to the performance outcomes in the table.

\begin{table}[h!]
\centering
\caption{Performance for declarative policies on one agent.}
\label{tab:policy_results}
\renewcommand{\arraystretch}{1.1}\setlength{\tabcolsep}{4pt}\footnotesize
\begin{tabular}{@{}lcccc@{}}
\toprule
\textbf{Policy} & \textbf{Time Windows} & \textbf{Waypoints} & \textbf{Edges} & \textbf{Reward} \\
\midrule
naive         & 7 & 112 & 160 & 1407.0 \\
front+        & 7 & 106 & 157 & 1664.0 \\
\textbf{poi\_focus} & \textbf{7} & \textbf{112} & \textbf{167} & \textbf{1725.0} \\
sanct\_soft   & 7 & 107 & 150 & 1490.0 \\
front+poi     & 7 & 106 & 150 & 1574.0 \\
\bottomrule
\end{tabular}
\end{table}

\section{Discussion}

Our experimental results highlight the primary strengths of the KG-based architecture by visually demonstrating the end-to-end planning process. The flow of the results section, from compiling a ``worldview'' to generating a multi-horizon plan provides qualitative evidence for the synergy between the two planes of our framework. Figures~\ref{fig:tensor_fusion} and~\ref{fig:agent_worldview} show the Data Plane providing the semantic ``what'' (mission value, constraints), while Figures~\ref{fig:graph_stitching} and~\ref{fig:local_path} show the Control Plane providing the physically-aware ``how'' (temporal stitching, current-aware navigation). This separation and recombination is the core of the system's strength: it allows for the generation of paths that are both scientifically valuable and physically efficient, a synergy that na\"{i}ve, value-only approaches (Fig.~\ref{fig:nav_graph_t0}) cannot achieve.

The policy differentiation experiment (Table~\ref{tab:policy_results} and Fig.~\ref{fig:global_and_local_plan}) is arguably the most compelling demonstration of our framework's value. The ability to dramatically alter agent behavior, from aggressive front-tracking to cautious sanctuary-avoidance, simply by changing declarative facts in a graph is a significant departure from conventional planning systems. This moves mission configuration from hard-coded cost functions or brittle scripts to a flexible, queryable, and auditable knowledge base. An operator can now ask ``why was this path chosen?'' and receive an answer rooted in traceable facts like \texttt{ex:usesPolicy ex:poi\_focus}. This represents a fundamental shift toward more explainable and adaptable autonomous systems.

This approach is not without its limitations. The performance of the system is heavily dependent on the quality and expressiveness of the KG's schema and the ingested data. Crafting a good ontology is a non-trivial task. Furthermore, while the KG enables fast, incremental replanning by identifying ``dirty'' windows, the initial compilation of all artifacts can be computationally intensive, though highly parallelizable.

Future work will focus on three areas. First, we will explore more sophisticated coordination strategies in the Control Plane, moving from greedy selection to market-based auction mechanisms for more optimal task allocation. Second, we aim to incorporate machine learning to automatically learn policy weights from mission outcomes, closing the loop from execution back to strategy. Finally, we will expand the ontology to include more complex concepts of uncertainty and risk, allowing the KG to reason about not just the most rewarding path, but the most robust one.

\section{Conclusion}

In this work, we presented a novel framework that uses a Knowledge Graph as an active translation layer for multi-agent path planning. By formally separating mission-specific semantics from a domain-agnostic planner, our approach bridges the critical gap between high-level objectives and low-level execution. The KG's two-plane architecture, a Data Plane that compiles mission-aware tensors and a Control Plane that provides physics-aware traversal rules enables the generation of intelligent, coordinated, and policy-driven plans.

Our experimental results visually demonstrated the end-to-end capability of this architecture in the complex, dynamic environment of the Gulf of Mexico. We showed the compilation of disparate semantic layers into a coherent agent ``worldview" and the subsequent generation of multi-horizon, temporally-stitched plans. Most critically, we proved that this architecture provides profound flexibility. The ability to specify and modify complex agent behaviors, from naïve exploration to targeted, constraint-aware operations, simply through declarative facts makes the system more adaptive, explainable, and robust. This work establishes the KG not merely as a data repository, but as a powerful, stateful orchestrator for autonomous systems.

\vspace{-0.25\baselineskip} 
\section*{Acknowledgment}
This work was partly supported by the Office of Naval Research (ONR) and the Naval Research Laboratory under contracts N0073-16-2-C902 and N00173-20-2-C007.

\vspace{-0.25\baselineskip} 

\small 
\bibliographystyle{IEEEtran}
\bibliography{references}

\end{document}